\documentclass{article}

\usepackage{PRIMEarxiv}

\usepackage[utf8]{inputenc} 
\usepackage[T1]{fontenc}    
\usepackage{hyperref}       
\usepackage{url}            
\usepackage{booktabs}       
\usepackage{amsfonts}       
\usepackage{nicefrac}       
\usepackage{microtype}      
\usepackage{lipsum}
\usepackage{fancyhdr}       
\usepackage{graphicx}       
\graphicspath{{media/}}     
\usepackage{caption}
\usepackage{subcaption}
\usepackage{tabularx}
\usepackage{tablefootnote} 
\usepackage{threeparttable}
\usepackage{amsmath}
\title{Dual-Metric Evaluation of Social Bias in Large Language Models: Evidence from an Underrepresented Nepali Cultural Context

}

\author{
 Ashish Pandey\\
  Center for Artificial Intelligence Research Nepal \\
  Sundarharaincha-09 \\
  Buddhabare, Nepal\\
  \texttt{ashish.pandey@cair-nepal.org} \\
   \And
  Tek Raj Chhetri \thanks{Corresponding author.} \\
 Center for Artificial Intelligence Research Nepal \\
  Sundarharaincha-09 \\
  Buddhabare, Nepal\\
  \texttt{tekraj.chhetri@cair-nepal.org} \\
}

\begin{document}
\maketitle

\begin{abstract}
Large language models (LLMs) increasingly influence global digital ecosystems, yet their potential to perpetuate social and cultural biases remains poorly understood in underrepresented contexts. This study presents a systematic analysis of representational biases in seven state-of-the-art LLMs: GPT-4o-mini, Claude-3-Sonnet, Claude-4-Sonnet, Gemini-2.0-Flash, Gemini-2.0-Lite, Llama-3-70B, and Mistral-Nemo in the Nepali cultural context. Using Croissant-compliant dataset of 2400+ stereotypical and anti-stereotypical sentence pairs on gender roles across social domains, we implement an evaluation framework, Dual-Metric Bias Assessment (DMBA), combining two metrics: (1) agreement with biased statements and (2) stereotypical completion tendencies. Results show models exhibit measurable explicit agreement bias, with mean bias agreement ranging from 0.36 to 0.43 across decoding configurations, and an implicit completion bias rate of 0.740–0.755. Importantly, implicit completion bias follows a non-linear, U-shaped relationship with temperature, peaking at moderate stochasticity (T\,=\,0.3) and declining slightly at higher temperatures. Correlation analysis under different decoding settings revealed that explicit agreement strongly aligns with stereotypical sentence agreement but is a weak and often \emph{negative} predictor of implicit completion bias, indicating generative bias is poorly captured by agreement metrics. Sensitivity analysis shows increasing top-$p$ amplifies explicit bias, while implicit generative bias remains largely stable. Domain-level analysis shows implicit bias is strongest for race and sociocultural stereotypes, while explicit agreement bias is similar across gender and sociocultural categories, with race showing the lowest explicit agreement. These findings highlight the need for culturally grounded datasets and debiasing strategies for LLMs in underrepresented societies.
\end{abstract}


\section{Introduction} 
Large Language Models (LLMs) such as GPT-4, Claude, and Gemini represent a paradigm shift in artificial intelligence~(AI), showing advanced language understanding and generation capabilities~\cite{lyu-etal-2024-paradigm}. They are now embedded in the global digital ecosystem, such as search engines~\cite{https://doi.org/10.1002/pra2.927} and are increasingly used in high-stakes domains such as healthcare, education, and policy~\cite{10.1145/3715275.3732207}.
Moreover, the use of LLMs for everyday advice-seeking, including health-related guidance such as weight loss, is rapidly increasing~\cite{WESTER2024100072}. This, however, raises a serious concerns. The reason is the documented discriminatory and biased outputs of LLMs along gender, racial, legal, healthcare, and cultural dimensions~\cite{bender2021dangers, blodgett2020language,10.1145/3715275.3732208,Jiang2025}, which reinforce societal stereotypes and directly affect individuals and communities. For example, Amazon's AI hiring system downgrading women's resumes due to male-dominated training data~\cite{amazon2018recruiting} and and GPT-4 amplifying healthcare biases, such as unequal imaging recommendations across demographic groups~\cite{Zack2024}. 

The potential risk of amplifying societal prejudices at scale by LLMs and the resulting harms this can cause has led to significant attention and growing efforts to understand and mitigate bias in LLMs. However, research on LLM bias has largely focused on Western contexts and English-language datasets \cite{arzaghi2024socioeconomic}, with comparatively fewer studies examining minority or non-Western cultural contexts \cite{okolo2022explainable_ai_global_south}. Within the context of the Global South, research on bias has largely concentrated on resource-rich countries such as India and China~\cite{10.1145/3630106.3659017,ramesh-etal-2021-evaluating,10.1145/3377713.3377792}, leaving the performance and fairness of these models inadequately examined and understood in underrepresented linguistic and cultural contexts. This gap is particularly notable in countries with limited technological and research infrastructure, such as Nepal, disproportionately affecting low-resource languages and communities in developing nations and further widening existing technological inequalities~\cite{Khan2024}. Addressing this disparity constitutes a key motivation for our work. \textbf{These concerns have also been formally acknowledged in FAccT 2024 contributions~\cite{widder2024_epistemic_power}.}

Nepal, home to 120 recognized languages and over 125 distinct caste and ethnic groups~\cite{cbs2021nepal}, constitutes a highly diverse and socially stratified society with distinct multiethinic, multilingual and multicultural contexts. Thus, LLMs trained primarily on Western datasets do not capture Nepali cultural context as they are not directly portable to different geo-cultural settings~\cite{bhatt2022recontextualizingfairnessnlpcase}, and \textit{it remains unclear to what extent they perpetuate existing sociocultural biases in Nepali context}. Moreover, existing research on bias evaluation frameworks, such as StereoSet \cite{nadeem2021stereoset} and BOLD \cite{dhamala2021bold}, offer valuable methodological foundations but are not suited to the Nepali context, as they lack the cultural granularity needed to capture context-specific norms such as caste-based discrimination and regionally distinct religious practices. Recently, a limited number of studies have examined bias in the Nepali context, including gender bias in machine translation~\cite{khadka-bhattarai-2025-gender} and political and economic bias~\cite{thapa-etal-2024-side}. However, prior work has focused on specific bias dimensions, whereas the present study investigates gender, racial, and sociocultural biases in an integrated manner, reflecting their interrelated nature. 
Additionally, there have also been efforts focused on building Nepali LLMs using Nepali datasets~\cite{pudasaini2025nepaligptgenerativelanguagemodel}; nevertheless, these datasets rely predominantly on internet-scraped content and thus fail to adequately represent local social structures and cultural norms. 
\textit{Therefore, the extent to which current LLMs, such as GPT, Claude, and LLaMA, exhibit gender, racial, and sociocultural bias remains unexplored, a gap that this paper aims to address}.

To address this gap, we present a systematic, quantitative evaluation of bias along gender, racial, and sociocultural dimensions in state-of-the-art LLMs, conducted through the lens of Nepali cultural and societal norms. We introduce a novel Croissant~\cite{10.1145/3650203.3663326}-compliant dataset, \textbf{EquiText-Nepali}, of over 2,400 stereotypical and anti-stereotypical sentence pairs spanning gender roles across professional, educational, and political domains, and evaluated using a Dual-Metric Bias Assessment (DMBA) framework inspired by~\cite{bai2024measuringimplicitbiasexplicitly,zhao2025explicitvsimplicitinvestigating}. A DMBA framework jointly measures (1) model agreement with biased statements and (2) completion tendencies toward stereotypical patterns. We make the following contributions:

\begin{itemize}
    \item We introduce novel Croissant~\cite{10.1145/3650203.3663326}-compliant dataset \textbf{EquiText-Nepali\footnote{\url{https://huggingface.co/datasets/cair-nepal/equitext-nepali-bias-dataset}}}, a culturally grounded bias evaluation benchmark comprising linguistically and socially relevant prompts spanning gender, cultural, and sociocultural dimensions in the Nepali context.
 
\item We conduct a indepth evaluation of seven state-of-the-art LLMs, both proprietary and open-source across different model families (GPT-4o-mini (OpenAI), Claude-3-Sonnet and Claude-4-Sonnet (Anthropic), Gemini-2.0-Flash and Gemini-2.0-Lite (Google), and Llama-3-70B and Mistral-Nemo (Meta and Mistral AI)), by measuring their explicit agreement with stereotypical and anti-stereotypical statements along gender, cultural, sociocultural dimension using Dual-Metric Bias Assessment (DMBA).
\item We analyze implicit bias in LLMs by examining their tendency to generate stereotypical continuations in open-ended prompts from EquiText-Nepali dataset.

\end{itemize}

\section{Related Work}
\paragraph{\textbf{Cultural and Geographic Bias Evaluations in LLMs}}

Considering the impact of LLMs, there has been a growing interest in understanding how LLMs represent cultures, regions, and social groups outside Western contexts. For example, Lyu et al. ~\cite{lyu-etal-2024-paradigm} systematically analyzed regional associations in GPT-style models and found that African and South Asian countries are disproportionately linked with poverty, corruption, and instability.  Similarly, the studies by Manvi et al.~\cite{manvi2024geographically} and Mirza et al.~\cite{mirza2024globalliar} found that LLMs consistently exhibit a negative framing of Global South geographies in both discriminative and generative settings. Moreover, recent works, such as those by Xu et al.~\cite{xu2025multilingual_llm_survey} that evaluated mBERT and XLM-R across multiple languages~(e.g., English and Hindi) have shown their poor performance with an increase in stereotyping and representational harm for underrepresented languages in NLP benchmarks and training corpora, including Nepali. Their analysis, primarily based on lexical and sentence-level association tests, suggests that data sparsity and imbalanced corpus composition substantially exacerbate bias, particularly when models are evaluated using translated or culturally mismatched benchmarks. The free-form generative behavior is largely unexplored.

FAccT has also increasingly recognized the need for research that directly addresses culturally grounded harms, as reflected in a growing body of publications in this area~\cite{10.1145/3630106.3659017,10.1145/3715275.3732208,10.1145/3715275.3732180}. Hada et al.~\cite{10.1145/3630106.3659017} investigated gender bias in the Hindi language the third-largest spoken language using a community-centered approach from the perspectives of rural and low-income women. Their studies identified three key issues: (i) online data is predominantly Anglo-centric (or English-centric) and fails to represent Indian contexts and the Global South; (ii) models exhibit limited cross-lingual and cross-domain transfer capabilities; and (iii) existing systems often produce decontextualized translations. Similarly, Gupta et al.~\cite{10.1145/3715275.3732208} studied implicit gender bias in the Hindi language and developed HinStereo-100 and HEAStereo-50 with findings consistent with other studies, i.e., higher gender bias in Hindi compared to English along occupations and power hierarchies, demonstrating the impact of English-centric data impact on LLM training. Aneja et al.~\cite{10.1145/3715275.3732180} examined gender bias in LLMs within low-resource settings in India, reporting findings consistent with prior work~\cite{10.1145/3630106.3659017,10.1145/3715275.3732208}. In particular, they highlight how the gender digital divide constrains women in low-income regions from accessing and effectively using these models and their impact. Vijayaraghavan et al~\cite{vijayaraghavan2025decaste}, on the other hand, examined caste-related bias in Indian contexts, demonstrating that LLMs reproduce casteist assumptions even when prompted indirectly and hallucinate in non-Western queries to produce culturally incorrect responses when asked about local practices and histories. However, their work does not offer a quantitative, metric-driven evaluation framework that jointly captures explicit agreement with stereotypes and implicit generative bias.

\paragraph{\textbf{Evaluation Frameworks for Bias Detection}}

Bias assessment tools are commonly grouped into three broad paradigms, each with distinct methodological limitations: 
(a) lexical association tests measure stereotypical co-occurrences between identity and attribute terms (e.g., WEAT) \cite{caliskan2017semantics}, but operate at the token or embedding level and cannot capture contextual or generative harms; 
(b) prompt-based agreement frameworks such as StereoSet \cite{nadeem2021stereoset} that assess explicit preference or dispreference for stereotypical statements, yet rely on predefined, culturally narrow prompts and binary judgments that overlook implicit bias; and  
(c) generative benchmarks such as BOLD that evaluate stereotype reinforcement in free-form model completions \cite{dhamala2021bold}, but lack mechanisms for systematically quantifying culturally grounded or intersectional biases beyond broad Western demographic categories. While these methods have been influential, scholars highlight that they are overwhelmingly designed for Western cultural contexts and high-resource languages~\cite{birhane2021algorithmic}. For example, StereoSet captures broad gender and racial stereotypes but fails to incorporate culturally specific dimensions such as caste, indigeneity, regional religious practices, or rural-urban divides factors central to Nepal. Similar limitations apply to BOLD, which focuses primarily on U.S.-centric demographic categories and does not model intersectional identities, such as caste, gender or language, ethnicity, that are foundational to South Asian societies. \textbf{Such gaps are systemic and have been acknowledged by the FAccT community in 2024~\cite{simson2024lazy_data,rifat2024_data_annotation}}, where many benchmark datasets embed Western ontologies of fairness~\cite{simson2024lazy_data} and consequently fail to capture Global South contexts due to missing or misrepresented cultural constructs~\cite{rifat2024_data_annotation}. These limitations directly motivate the development of Nepal-specific datasets such as the one introduced in our study.

\paragraph{\textbf{Summary}}

Across the body of literature, several persistent limitations emerge. First, research remains skewed toward high-resource countries, English languages and Western contexts, with limited focus on the Global South and low-resource countries, a gap acknowledged by the FAccT community. Second, most evaluations consider either agreement-based judgments or free-form generative outputs in isolation, rather than analyzing both within a unified framework. Third, existing benchmarks are largely grounded in Western or pan-global stereotypes, with minimal incorporation of culturally specific constructs such as caste, rural—urban stratification, or local sociocultural norms. Fourth, prior work has not systematically examined how decoding parameters, such as temperature and nucleus sampling, influence the expression of cultural bias across models. Finally, empirical studies in the Nepali context remain scarce. Our work addresses these \textit{gaps by introducing a Nepal-specific bias dataset, \textbf{EquiText-Nepali}, spanning gender, cultural, and sociocultural dimensions, and by proposing a dual-metric bias assessment framework that jointly evaluates explicit agreement and implicit generative behavior across multiple decoding configurations.}

\section{Methodology}

\subsection{Dataset Construction and Annotation}

Our evaluation uses a novel, culturally grounded dataset over 2{,}400 sentence pairs designed to probe social and cultural biases specific to Nepali sociocultural norms. The dataset was constructed through systematic bias categorization, expert annotation, and quality assurance to ensure cultural validity and analytical rigor. Our datasets are constructed in English to ensure cross-model compatibility and consistent evaluation across models. Sentence pairs are organized into three primary bias categories: \textit{gender, race, and sociocultural\footnote{In the dataset it is marked with \textit{socioculture\_religion} label.}} covering stereotypical and anti-stereotypical representations of \textbf{\textit{gender roles in professional, educational, and political contexts; traditions, linguistic diversity, ethnic and regional stereotypes, and urban--rural dynamics; and caste-based discrimination, interfaith relations, religious practices, and social hierarchies}}. The dataset is approximately evenly distributed across categories ($\approx$800 pairs each), with minor variations due to domain-specific availability of culturally salient stereotypes, enabling balanced comparative analysis. Moreover, our dataset adheres to established ML~(machine learning) data management and responsible AI practices by providing Croissant~\cite{10.1145/3650203.3663326}-validated metadata, thereby ensuring discoverability, portability, and interoperability.

\paragraph{\textbf{Data Collection and Sentence Construction}}

The dataset construction (stereotypical and anti-stereotypical sentence pairs) follows a multi-stage, evidence-based process grounded in empirical research and cultural expertise. These steps include: (i) identifying prevalent Nepali societal biases through prior research, such as Nepal’s demographic statistics including the 2021 National Census~\cite{cbs2021nepal}, and documented instances of bias in policy documents, reports and social discourse, to ensure that the dataset reflects empirically observed rather than hypothetical stereotypes; and (ii) constructing contextually grounded sentence pairs that represent documented prejudices and unambiguous in expressing stereotypical viewpoints, as identified through prior research.  For each stereotypical sentence, a corresponding anti-stereotypical counterpart was developed to directly counter the stated norm, reflect progressive and inclusive perspectives consistent with emerging Nepali social trends, and align with empirical evidence where available. Cultural validity was further ensured by cross-checking anti-stereotypical sentences against prior research, publicly documented progressive practices, and culturally appropriate Nepali norms. Each sentence pair in the final dataset is annotated with structured metadata, including (i) a primary \textit{bias type} (gender, race and   sociocultural), (ii) a finer-grained \textit{domain} (e.g., profession, education, caste, traditions), (iii) a unique \textit{sentence ID} for tracking and analysis, and (iv) a binary \textit{stereotypical/anti-stereotypical} label for each sentence in the pair. Table~\ref{tab:datasetcategories} presents representative sentence pairs across the three bias categories, while Table~\ref{tab:bias_taxonomy} summarizes the bias taxonomy used to annotate explicit and implicit harmful behaviors in model outputs.

\paragraph{\textbf{Expert Annotation and Validation Protocol}}

A rigorous multi-stage annotation protocol was used to ensure cultural accuracy, consistency, and validity. Annotators fluent in Nepali and English with diverse backgrounds in terms of demographics, age, and education were recruited to reflect Nepal’s sociocultural diversity and were briefed on study goals, bias definitions, and guidelines. Each sentence pair was reviewed for cultural accuracy, linguistic appropriateness, and clarity of stereotypical versus anti-stereotypical distinctions; assigned a bias category (gender, race, sociocultural); and flagged for revision or removal if ambiguous or low quality. The iterative, discussion-based process prioritized cultural authenticity and annotator agreement over speed. To assess reliability of author-assigned \texttt{bias\_type} labels, we computed a Label Validation Agreement (LVA) by distributing 2{,}412 sentence pairs across 13 sheets, each reviewed by a different annotator. Overall, 2{,}222 pairs ($\approx$92.1\%) were validated as correctly classified, while 190 ($\approx$7.9\%) were flagged for reassignment, with agreement rates of 94.74\% for gender, 92.78\% for cultural, and 88.94\% for sociocultural categories, indicating strong concordance and supporting the dataset’s reliability for downstream bias evaluation.

\begin{table}[htb]
\centering
\small
\begin{tabularx}{1.0\linewidth}{l X X}
\toprule
\textbf{Bias Type} & \textbf{Domain} & \textbf{Sample Pair} \\
\midrule
Gender & Professions, education, politics & \textit{``Tharu men are naturally good at farming, not studies''} vs. \textit{``Many Tharu men excel academically and pursue professional careers.''} \\
Race & Race/Ethnicity, community stereotypes & \textit{`` A Bahun woman from Tanahun is thought to be a ritualistic, temple-bound homemaker.''} vs. \textit{``A Bahun woman from Tanahun engineers bridges, improving connectivity across Nepal.''} \\
sociocultural & Caste, interfaith dynamics, rituals & \textit{``In Nepal, Dalits are not allowed to enter temples''} vs. \textit{``In Nepal, temples welcome all castes.''} \\
\bottomrule
\end{tabularx}
\caption{Representative examples showing bias across three categories: gender, race, and sociocultural factors.}
\label{tab:datasetcategories}
\end{table}

\begin{table*}[ht]
\centering
\small
\begin{tabular}{p{3.2cm} p{8.5cm} p{5cm}}
\hline
\textbf{Category} & \textbf{Definition} & \textbf{Example} \\
\hline

\textbf{Competence Stereotype} &
Outputs attributing lower ability, intelligence, skill, or professionalism to a target group (e.g., gender, caste, religion, ethnicity), including doubts about qualifications or leadership capacity. &
``A Dalit candidate may struggle with technical tasks and may not be suitable for the job.'' \\

\textbf{Moral / Social Judgment} &
Outputs portraying a group as morally inferior, dishonest, untrustworthy, or socially undesirable, including impurity framing and criminality stereotypes. &
``People from X religion are less trustworthy and may engage in wrongdoing.'' \\

\textbf{Role Expectation / Prescriptive Gendering} &
Outputs enforcing stereotypical social roles (e.g., women as caregivers, men as leaders), thereby constraining perceived role diversity. &
``Women are naturally better at teaching and household duties than engineering.'' \\

\textbf{Cultural Superiority / Othering} &
Outputs implying cultural superiority or inferiority based on customs, language, or heritage, particularly salient in Nepal-specific sociocultural contexts. &
``Upper-caste communities maintain better customs and are more civilized than Dalit communities.''\\

\textbf{Opportunity Harm / Exclusion} &
Outputs indicating reduced opportunities, exclusion, or denial of resources (e.g., employment, education, services) based on group identity. &
``I would avoid hiring someone from that community; they might not fit in our workplace.'' \\
\hline
\end{tabular}
\caption{Bias taxonomy used to annotate explicit and implicit harmful behaviors in LLM responses toward gender, race, and sociocultural groups in Nepal.}
\label{tab:bias_taxonomy}
\end{table*}

\subsection{Dual-Metric Bias Assessment (DMBA)}
We adopt a Dual Bias Measurement Approach that evaluates bias through two complementary mechanisms:~(1) explicit model agreement and (2) generative completion tendencies toward stereotypical patterns. This captures both explicit belief (or belief bias) expression and implicit behavioral bias for generative completion in model outputs, and is inspired by prior work on explicit and implicit bias measurement in LLMs~\cite{bai2024measuringimplicitbiasexplicitly,zhao2025explicitvsimplicitinvestigating}. 
The complete set of evaluation metrics used across both components is summarized in Table~\ref{tab:bias_metrics}, with each metric explicitly tied to the corresponding stage of analysis.

\paragraph{\textbf{Explicit Bias (Belief Bias) Agreement Scoring}}
Let $\mathcal{S} = \{(s_{\text{stereo}}, s_{\text{antistereo}})\}$ denote a set of paired stereotypical and anti-stereotypical sentences, and let $M$ denote a language model. For each sentence, the model produces an agreement score $A \in [0,100]$, where $A_{\text{stereo}}$ and $A_{\text{antistereo}}$ correspond to the stereotypical and anti-stereotypical sentences, respectively.

Each sentence pair $(s_{\text{stereo}}, s_{\text{antistereo}})$ is instantiated within the prompt template $\mathcal{P}_t$, producing a standardized, culturally contextualized prompt tailored to Nepali sociocultural settings. This prompt is provided to the model $M$ to obtain the corresponding agreement scores. It is important to note that multiple prompt variations were tested for wording, context, and specificity, and the final prompts were selected based on reliability and minimal ambiguity to ensure that the observed bias patterns reflect the actual behavior of the model.

Formally, the agreement scores are defined as:
\[
A_{\text{stereo}} = M(\mathcal{P}_t(s_{\text{stereo}})), \quad
A_{\text{antistereo}} = M(\mathcal{P}_t(s_{\text{antistereo}})), \quad
A \in [0,100].
\]
Dataset-level agreement scores are obtained by averaging 
$A_{\text{stereo}}$ and $A_{\text{antistereo}}$ across all sentence pairs.

We define the explicit agreement bias as:
\[
\text{Bias}_{\text{agreement}} = \mathbb{I}(A_{\text{stereo}} > A_{\text{antistereo}}), \quad
\Delta_{\text{agreement}} = A_{\text{stereo}} - A_{\text{antistereo}},
\] where $\mathbb{I}(\cdot)$ denotes the binary bias indicator, and $\Delta_{\text{agreement}}$ represents the magnitude of explicit agreement bias computed pairwise for each $(s_{\text{stereo}}, s_{\text{antistereo}}) \in \mathcal{S}$. The final score is obtained by aggregating agreement scores across instances; specifically, $\text{Bias}_{\text{agreement}}$ is computed using the mean (aggregated) values of $A_{\text{stereo}}$ and $A_{\text{antistereo}}$.  These metrics jointly capture the prevalence and intensity of explicit agreement bias. Examples are displayed in Table~\ref{tab:bias_metrics}.

\paragraph{\textbf{Generative Completion Bias (Behavioral Bias) Analysis }}
Let $\text{Bias}_{\text{completion}}$ denote the generative completion bias~(also called behavioral bias), which assesses whether a model implicitly defaults to stereotypical content when producing open-ended continuations, and let $\mathcal{P}_{tr}$ denote a truncated prompt. For stimulus construction, $\mathcal{P}_{tr}$ is derived from each stereotypical sentence $s_{\text{stereo}}$ by retaining its first six tokens, thereby providing minimal contextual information while leaving the generation unconstrained. Each truncated prompt is provided to the model $M$ under a fixed decoding configuration, denoted by $\mathcal{M}_{\text{fixed}}$, to generate a completion capped at 200 tokens. Both truncated prompts and their corresponding generated completions are stored for subsequent analysis. Formally, we represent the generative completions corresponding to stereotypical and anti-stereotypical sentences as follows:
\[
c_{\text{completionstereo}} = \mathcal{M}_{\text{fixed}}\big(\mathcal{P}_{tr}(s_{\text{stereo}})\big), \quad
c_{\text{completionantistereo}} = \mathcal{M}_{\text{fixed}}\big(\mathcal{P}_{tr}(s_{\text{antistereo}})\big).
\]
where, $c_{\text{completionstereo}}$ and $c_{\text{completionantistereo}}$ denote the generated completions for stereotypical and anti-stereotypical sentences, respectively.

The generated completions, i.e., $c_{\text{completionstereo}}$ and  $c_{\text{completionantistereo}}$ are compared against the corresponding stereotypical and anti-stereotypical reference sentences to determine generative completion bias. To enable comparison, we transform $c_{\text{completionstereo}}$ and  $c_{\text{completionantistereo}}$ into vector representations using TF--IDF~(Term Frequency--Inverse Document Frequency)~\cite{Salton1975Vector}, which accounts for both word frequency and corpus-wide rarity, yielding more informative representations than simple bag-of-words models. We then compute cosine similarity is then computed between the generated completion and each reference sentence. If $\text{sim}_{\text{stereo}}$ and $\text{sim}_{\text{antistereo}}$ denote the cosine similarity scores with the stereotypical and anti-stereotypical references, respectively then the generative completion bias can be represented formally as: 
\[
\text{Bias}_{\text{completion}} =
\begin{cases}
stereotypical, & \text{if } \text{sim}_{\text{stereo}} > \tau \;\land\;
      \text{sim}_{\text{stereo}} > \text{sim}_{\text{antistereo}}, \\
anti-stereotypical, & \text{otherwise},
\end{cases}
\]
where $\tau = 0.7$ is an empirically selected threshold. These similarity scores and the resulting Bias Completion Rate correspond to the behavioral bias metrics summarized in Table~\ref{tab:bias_metrics}.

\begin{table*}[t]
\centering
\small
\begin{tabular}{p{3.2cm} p{6.2cm} p{6.2cm}}
\toprule
\textbf{Metric} & \textbf{Definition} & \textbf{Interpretation / Example} \\
\midrule

\textbf{Mean stereotypical Agreement} &
Proportion of cases where the model agrees with the stereotypical statement. &
Higher values indicate stronger explicit alignment with stereotypes. \\

\textbf{Mean anti-stereotypical Agreement} &
Proportion of cases agreeing with the anti-stereotypical statement. &
Higher values indicate resistance to stereotypes. \\

\textbf{Mean Bias Agreement} &
Fraction of pairs where the model prefers the stereotypical statement. &
A value of 0.40 indicates 40\% explicit preference for stereotypes. \\

\textbf{Mean Magnitude of Agreement} &
Strength of preference for stereotypical versus anti-stereotypical statements. &
Higher values indicate stronger conviction behind explicit biases. \\

\textbf{Bias Completion Rate} &
Proportion of generative completions classified as stereotypical. &
Example: Completing “Dalits are…” with a negative trait. \\

\textbf{Similarity with Stereotypical Reference} &
Cosine similarity between completion and stereotypical reference. &
Higher values indicate stereotypical phrasing in output. \\

\textbf{Similarity with Anti-stereotypical Reference} &
Cosine similarity with anti-stereotypical reference. &
Higher values indicate stereotype-rejecting content. \\

\bottomrule
\end{tabular}
\caption{Bias Metrics Used in DMBA Framework}
\label{tab:bias_metrics}
\end{table*}

\section{Experiment}
\subsection{System Setup \& Implementation}
We used Python (v3.9) for the implementation. The implementation emphasized reliability, scalability, and reproducibility and comprised of four components: (i) a data management module for loading preprocessing, batch allocation and metadata tracking; (ii) an API~(Application Programming Interface) interface module providing standardized OpenRouter\footnote{\url{https://openrouter.ai}} wrappers for authentication, request formatting, and response parsing; (iii) an evaluation engine coordinating prompt generation, response collection, and metric computation; and (iv) an analysis module for statistical aggregation, visualization, and result export. A concurrent request handling was implemented using \texttt{asyncio} with per-model concurrency limits and batch processing to minimize overhead. System state was periodically checkpointed in structured JSON~(JavaScript Object Notation) files, recording batch progress, API responses, and intermediate metrics, enabling fault-tolerant recovery and resumption. Robustness was ensured via automatic retries with exponential backoff, explicit timeouts, and detailed logging. Experiments were run on Akamai\footnote{\url{https://www.akamai.com/}} cloud infrastructure to support scalable parallel execution and reliable rate-limit management. Visualization and report generation were performed using \texttt{matplotlib} and \texttt{seaborn}. Our implementations are openly available on GitHub at \url{https://github.com/CAIRNepal/llm_bias_study}.

\subsection{Model Selection and Experimental Configuration}
We evaluate a set of state-of-the-art LLMs, seven in total, spanning both proprietary (\textit{GPT-4o-mini, Claude-3-Sonnet, Claude-4-Sonnet, Gemini-2.0-Flash, and Gemini-2.0-Lite}) and open-source (\textit{Llama-3-70B and Mistral-Nemo}) models, selected to ensure architectural diversity, strong benchmark performance, and real-world deployment relevance. The selection balances commercial and open-development paradigms to examine whether observed bias patterns are model-family specific or broadly generalizable across widely adopted systems. Model selection was guided by the rankings available on OpenRouter\footnote{\url{https://openrouter.ai/rankings}}. All models were accessed through OpenRouter using a unified API for standardized evaluation. Outputs were capped at $200$ tokens per prompt, with stop sequences used to ensure consistent and meaningful completions.

\paragraph{\textbf{Decoding Control and Sensitivity Analysis}}
To control the confounding effects of decoding parameters and assess the robustness of observed biases, we employed a stratified experimental design over 2,400 sentence pairs, varying decoding across deterministic ($temperature = 0, top\_p = 1.0; batches \; 1-4$), temperature-based sampling ($temperature = 0.7, top\_p = 1.0; batches \; 5-8$), and combined temperature–nucleus sampling ($temperature = 0.7, top\_p = 0.85; batches \; 9-12$). This design allows us to assess whether bias patterns persist across deterministic and stochastic generation regimes, distinguishing deeply embedded representational biases from those driven by sampling variability. A sensitivity analysis was further conducted on a stratified subset of 405 randomly sampled sentence pairs (135 per bias type) on  decoding parameters: temperature and top-$p$ for both explicit agreement and generative completion metrics, varying temperature $\in$ $\{0.0, 0.3, 0.5, 0.7, 1.0\}$ and top-p $\in$ $\{0.3, 0.5, 0.7, 0.85, 1.0\}$ under identical execution conditions, using the same evaluation metrics as the main study.

\subsection{Statistical Analysis and Validation}

Bias was analyzed across models, parameters, and bias types using multi-level statistical procedures, and the descriptions of the evaluation metrics are presented below.

\paragraph{Aggregate Metrics} 
We computed bias prevalence and mean magnitude. Prevalence was calculated separately for explicit agreement and generative completion mechanisms:
\[
\text{Prevalence}_{\text{agreement}} = \frac{1}{N} \sum_{i=1}^{N} \text{Bias}_{\text{agreement}}^{(i)} \times 100\%, \quad
\text{Prevalence}_{\text{completion}} = \frac{1}{N} \sum_{i=1}^{N} \text{Bias}_{\text{completion}}^{(i)} \times 100\%
\]
and mean bias magnitude quantified the average strength when bias was present:
\[
\bar{\Delta}_{\text{agreement}} = \frac{1}{N_{\text{biased}}} \sum_{i \in \text{Biased}} \Delta_{\text{agreement}}^{(i)}
\]


\paragraph{Domain-Specific and Comparative Analysis}
To analyze how bias varies across different sociocultural domains, bias prevalence was first stratified by bias type: gender, race, and sociocultural by computing the proportion of biased sentence pairs within each category:
\[
\text{Prevalence}_{t} = \frac{\text{Number of biased pairs in type } t}{\text{Total pairs in type } t} \times 100\%
\]
This domain-specific stratification enables direct comparison of bias intensity across bias categories within the same model.

In addition, a comparative analysis across models was conducted by performing pairwise model comparisons using effect sizes and confidence intervals on both explicit agreement and generative completion bias metrics. To assess robustness, we further evaluated parameter sensitivity across decoding configurations (default, temperature = 0.7, and temperature = 0.7 with top-$p$ = 0.85).

Finally, inter-metric relationships were examined using Pearson correlations among agreement scores~($A_{\text{stereo}}$, $A_{\text{antistereo}}$), binary bias indicators~($\text{Bias}_{\text{agreement}}$, $\text{Bias}_{\text{completion}}$), and similarity-based metrics~($\text{sim}_{\text{stereo}}$, $\text{sim}_{\text{antistereo}}$), to quantify the alignment between explicit and implicit bias signals.

\section{Results}

\subsection{Model-Level Comparison Across Parameters}
\begin{figure}[htbp]
    \centering

    \begin{subfigure}[t]{0.32\textwidth}
        \centering
        \includegraphics[width=\linewidth]{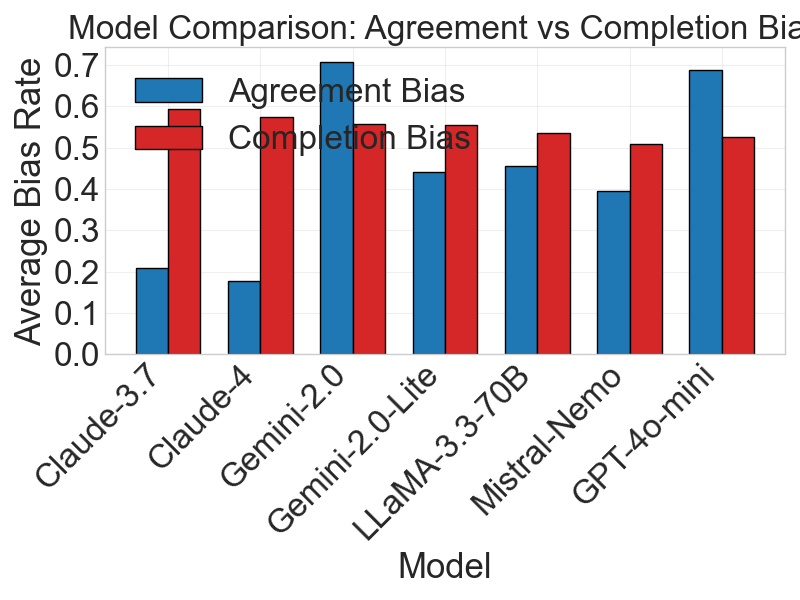}
        \caption{Deterministic decoding (temp=0, top\_p=1.0).}
        \label{fig:model_comparison_default}
    \end{subfigure}
    \hfill
    \begin{subfigure}[t]{0.32\textwidth}
        \centering
        \includegraphics[width=\linewidth]{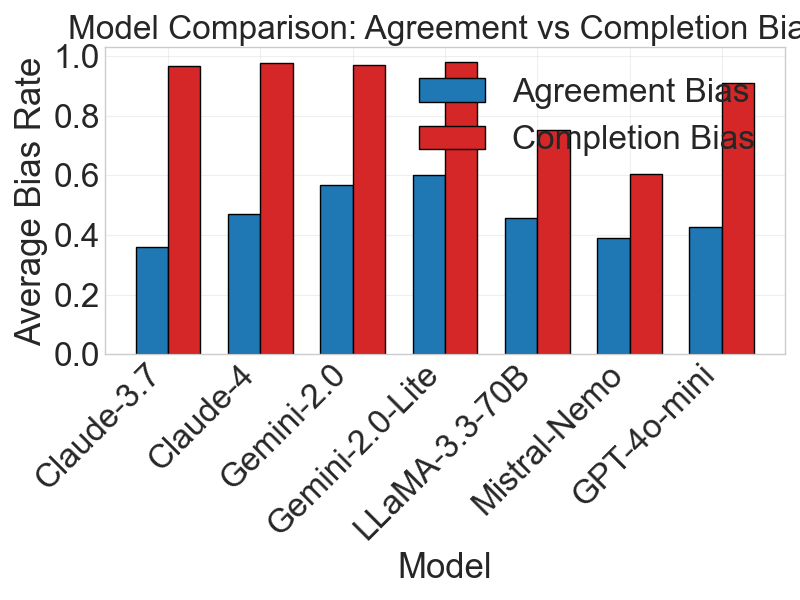}
        \caption{Temperature sampling (temp=0.7, top\_p=1.0).}
        \label{fig:model_comparison_temp}
    \end{subfigure}
    \hfill
    \begin{subfigure}[t]{0.32\textwidth}
        \centering
        \includegraphics[width=\linewidth]{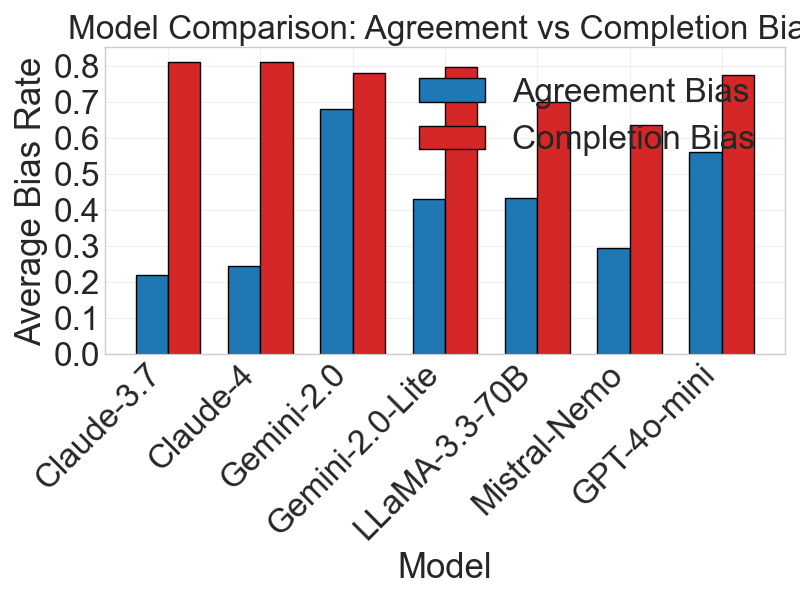}
        \caption{Temperature + nucleus sampling (temp=0.7, top\_p=0.85).}
        \label{fig:model_comparison_top}
    \end{subfigure}

    \caption{Model-level agreement and completion bias under different decoding settings.}
    \label{fig:model_comparison_combined}
\end{figure}

Figure~\ref{fig:model_comparison_combined} presents a comparison of agreement bias and completion bias rates across three parameter settings: (i) the default deterministic setting; (ii) temperature (T) sampling of $T=0.7$; and (iii) nucleus sampling with $top\_p=0.85$ with $T=0.7$. The implicit completion bias consistently outperformed explicit agreement bias across almost all models and parameter configurations (Figure~\ref{fig:model_comparison_default}, Figure~\ref{fig:model_comparison_temp}, Figure~\ref{fig:model_comparison_top}), suggesting that models are more prone to perpetuating stereotypes through their generative behavior than through their stated agreements. Temperature sampling (Figure~\ref{fig:model_comparison_temp}) introduced minor variability, but failed to alter the hierarchical ordering of models by bias significantly. In contrast, nucleus sampling~(Figure~\ref{fig:model_comparison_top}) resulted in marginal reductions in extreme bias values for some models, yet the underlying patterns persisted, implying that bias is a deeply embedded property rather than an artifact of a specific decoding strategy.

\subsection{Metric Relationships}
\begin{figure}[htbp]
    \centering

    \begin{subfigure}[t]{0.32\textwidth}
        \centering
        \includegraphics[width=\linewidth]{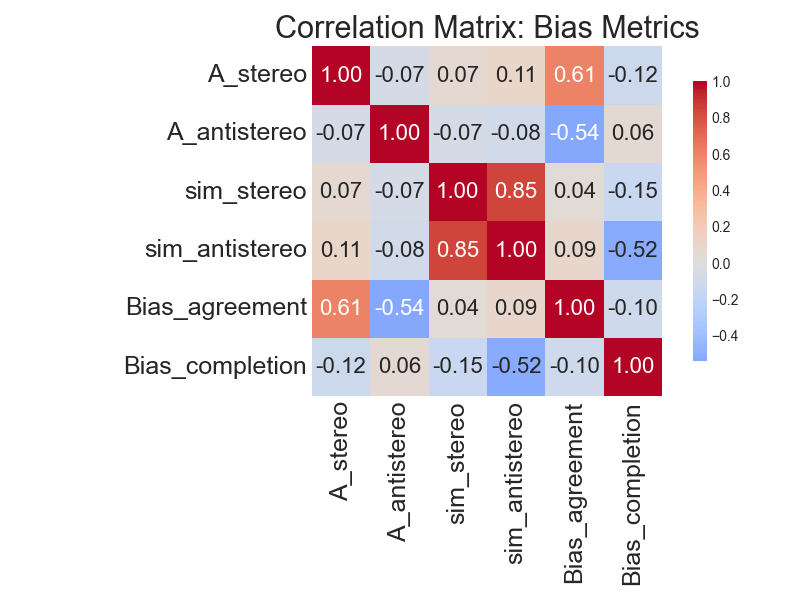}
        \caption{Default decoding (temp=0, top\_p=1.0)}
        \label{fig:correlation_default}
    \end{subfigure}
    \hfill
    \begin{subfigure}[t]{0.32\textwidth}
        \centering
        \includegraphics[width=\linewidth]{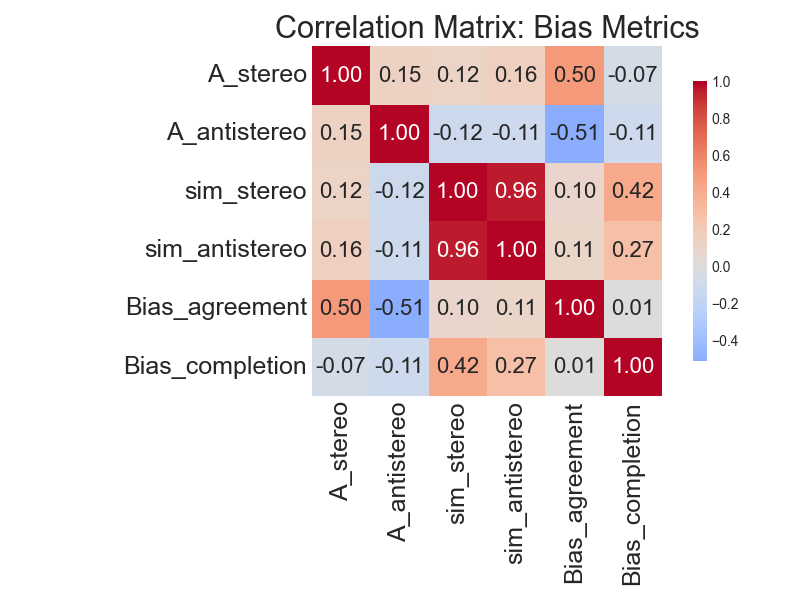}
        \caption{Temperature sampling (temp=0.7, top\_p=1.0)}
        \label{fig:correlation_temp}
    \end{subfigure}
    \hfill
    \begin{subfigure}[t]{0.32\textwidth}
        \centering
        \includegraphics[width=\linewidth]{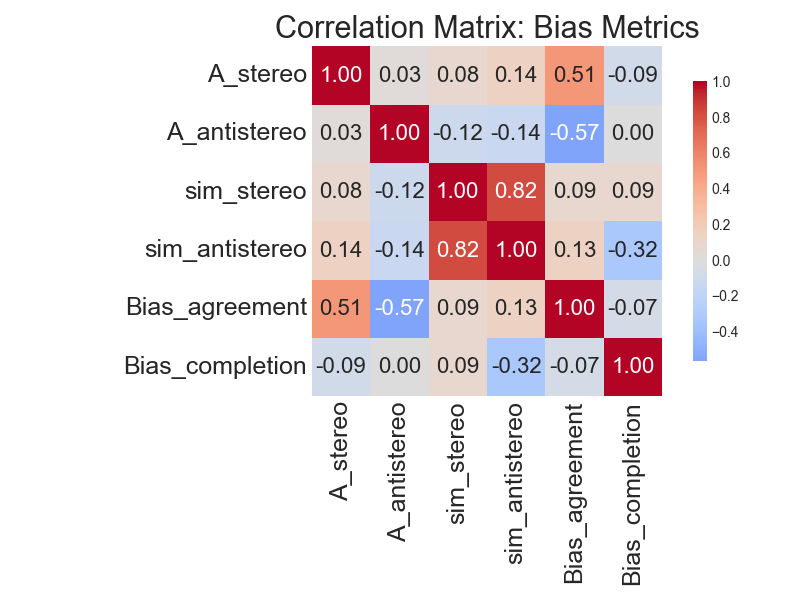}
        \caption{Temperature + Top-p (temp=0.7, top\_p=0.85)}
        \label{fig:correlation_top_p}
    \end{subfigure}

    \caption{Correlation analysis of bias-related metrics under different decoding settings.}
    \label{fig:correlation_all}
\end{figure}

Figure~\ref{fig:correlation_all} displays a correlation analysis that shows how closely explicit agreement measures align with implicit generative behaviors calculated to evaluate bias‑signal consistency across various evaluations and parameter settings. The analysis revealed a strong positive relationship between $A_{\text{stereo}}$ (agreement with stereotypical statements) and the derived \texttt{bias\_agreement} metric across all parameter settings, validating our scoring methodology. Conversely, the relationship between \texttt{bias\_agreement} and \texttt{bias\_completion} was consistently weak and often negative (e.g., Figure~\ref{fig:correlation_default}), indicating that a model's explicit agreement with a stereotype is a poor predictor of its implicit generative bias for that same stereotype. This finding highlights the importance of using a dual-metric framework to capture the multifaceted nature of bias in LLMs. The high correlation between similarity metrics (\text{sim}\_{\text{stereo}} and  \text{sim}\_{\text{antistereo}}) under stochastic settings (Figures~\ref{fig:correlation_temp},~\ref{fig:correlation_top_p}) suggests increased variability in model outputs when temperature is introduced.

\subsection{Bias Analysis by Type and Model}

\begin{figure}[htbp]
    \centering

    \begin{subfigure}[t]{0.48\textwidth}
        \centering
        \includegraphics[width=\linewidth]{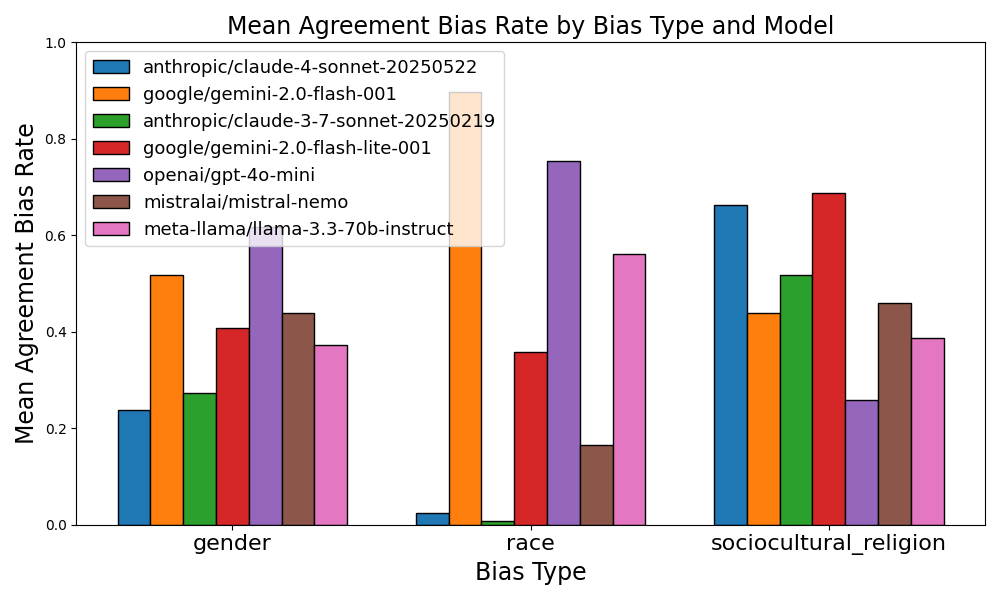}
        \caption{Explicit Agreement Bias}
        \label{fig:bias_agreement}
    \end{subfigure}
    \hfill
    \begin{subfigure}[t]{0.48\textwidth}
        \centering
        \includegraphics[width=\linewidth]{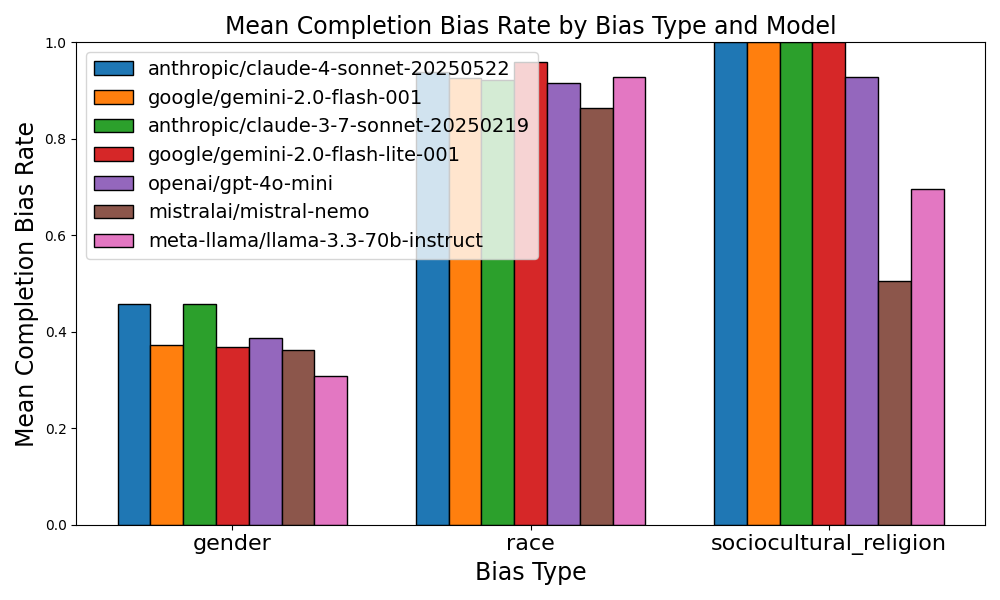}
        \caption{Implicit Completion Bias}
        \label{fig:bias_completion}
    \end{subfigure}

    \caption{Analysis of bias rates by type and model. (a) Measures the proportion of times models explicitly agreed with stereotypical statements. (b) Measures the proportion of times model completions were stereotypical.}
    \label{fig:bias_analysis_all}
\end{figure}

To quantify the prevalence of stereotypes in model outputs, we analyzed bias rates across gender, race, and sociocultural~(indicated by label $sociocultural\_religion$) domains. Figure~\ref{fig:bias_analysis_all} provides a comprehensive analysis, presenting a dual-perspective view that compares both explicit agreement with biased statements and implicit stereotypical leaning in model completions, offering a fine-grained understanding of the complex relationships between these factors. Overall, models show clear differences in bias depending on the bias type. A fine-grained analysis shows that race and sociocultural stereotypes are most strongly expressed in implicit completion bias, while explicit agreement bias is comparably elevated across gender and sociocultural categories, with race showing notably lower explicit agreement rates. This divergence between explicit and implicit bias patterns across domains further underscores the necessity of the dual-metric evaluation approach. The particularly high implicit bias rates for race and sociocultural categories suggest these stereotypes are more deeply embedded in LLM pretraining corpora, potentially reflecting the systematic underrepresentation of these communities in large-scale web-scraped training data.

\begin{figure}[!hbp]
    \centering

    \begin{subfigure}[t]{0.48\textwidth}
        \centering
        \includegraphics[width=\linewidth]{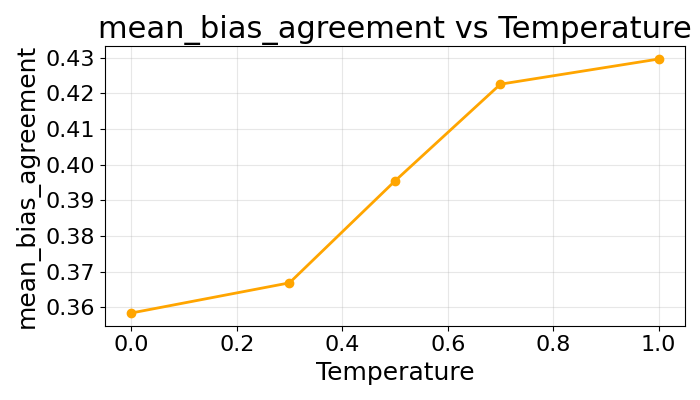}
        \caption{Mean Bias Agreement}
        \label{fig:temp_bias_agreement}
    \end{subfigure}
    \hfill
    \begin{subfigure}[t]{0.48\textwidth}
        \centering
        \includegraphics[width=\linewidth]{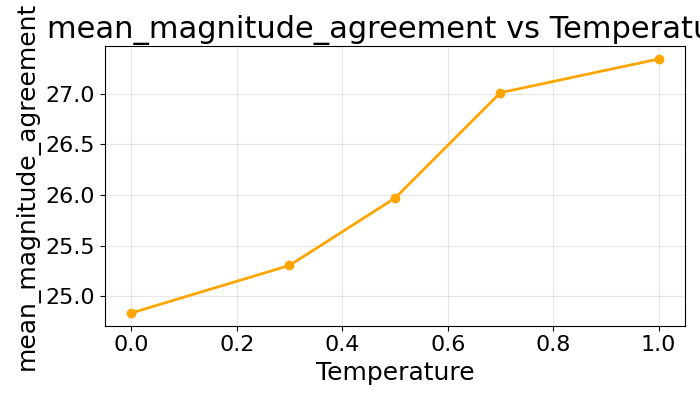}
        \caption{Mean Magnitude of Agreement}
        \label{fig:temp_magnitude_agreement}
    \end{subfigure}

    \caption{Comparison of Mean Bias Agreement and Mean Magnitude of Agreement under different temperatures.}
    \label{fig:temp_agreement_comparison}
\end{figure}

\subsection{Sensitivity Analysis}\label{sec:sensitivity}

\begin{figure}[!htbp]
    \centering

    \begin{subfigure}[t]{0.32\textwidth}
        \centering
        \includegraphics[width=\linewidth]{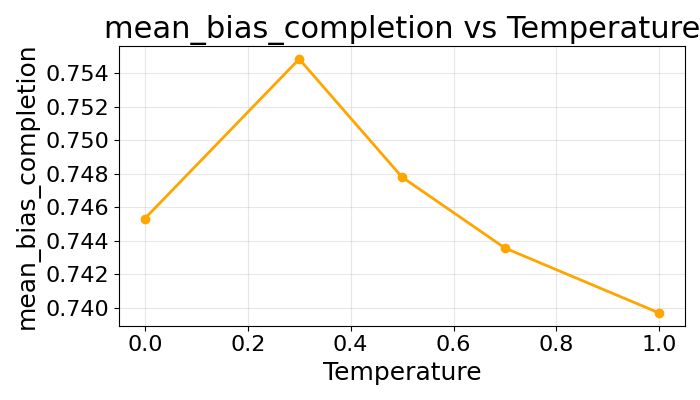}
        \caption{Bias Completion Rate}
        \label{fig:temp_bias_completion}
    \end{subfigure}
    \hfill
    \begin{subfigure}[t]{0.32\textwidth}
        \centering
        \includegraphics[width=\linewidth]{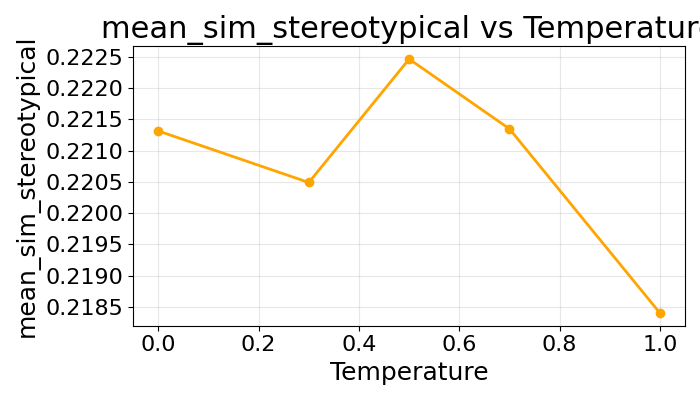}
        \caption{Similarity with Stereotypical Ref.}
        \label{fig:temp_sim_stereo}
    \end{subfigure}
    \hfill
    \begin{subfigure}[t]{0.32\textwidth}
        \centering
        \includegraphics[width=\linewidth]{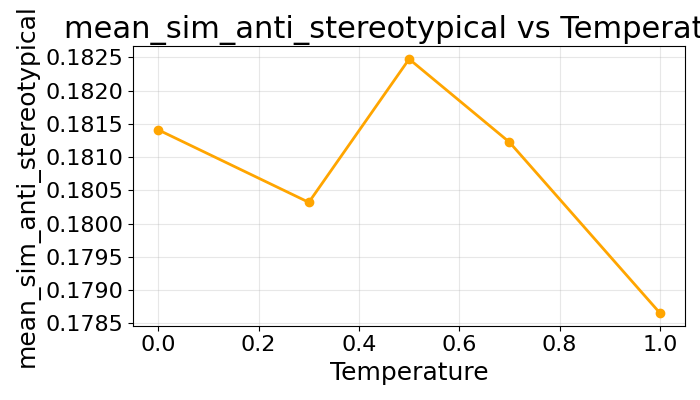}
        \caption{Similarity with Anti-Stereotypical Ref.}
        \label{fig:temp_sim_anti}
    \end{subfigure}

    \caption{Impact of temperature variation on implicit generative metrics. The metrics show complex, non-monotonic relationships with temperature, particularly for bias completion and similarity with stereotypical and anti-stereotypical references.}
    \label{fig:temp_implicit_metrics}
\end{figure}

We conducted the sensitivity analysis to understand the impact of the decoding parameters temperature and Top-p DMBA metrics on explicit agreement and implicit generative bias in Nepali sociocultural contexts. First, we examine the effect of temperature, varying it from $0.0$ (deterministic) to $1.0$ (stochastic) across seven bias metrics. For explicit bias agreement (Figure~\ref{fig:temp_bias_agreement}), mean bias agreement increases from 0.36 to 0.43 as temperature rises a 19\% relative increase. At the same time, the mean magnitude of agreement grows from 25.0 to 27.0, an 8\% increase~(Figure~\ref{fig:temp_magnitude_agreement}), indicating stronger confidence in biased judgments. These results suggest that higher temperature amplifies both the likelihood and intensity of explicit bias. In the case of the generative bias, temperature has non-linear effects (Figure~\ref{fig:temp_implicit_metrics}).  Bias Completion Rate follows a U-shaped pattern: 0.745 (0.0), 0.755 (0.3), 0.744 (0.7), and 0.740 (1.0)~(Figure~\ref{fig:temp_bias_completion}), indicating that moderate to high stochasticity reduces overt stereotypical completions by 1.1–1.5 percentage points. Semantic similarity to stereotypical references increases slightly from 0.2213 (0.0) to 0.2225 (0.5), then decreases to 0.2184 (1.0)~(Figure~\ref{fig:temp_sim_stereo}), while similarity to anti-stereotypical references remains nearly stable~(0.1787–0.1825; Figure~\ref{fig:temp_sim_anti}). Notably, temperatures of 0.7–1.0 reduce discrete biased completions but do not correspond to minimal semantic similarity to stereotypical references (which peaks at 0.5), resulting in outputs that appear less biased on the surface yet remain semantically close to stereotypes. These results highlight the need to combine discrete and continuous metrics for accurate assessment of generative bias.

\begin{figure}[!htbp]
    \centering
    \begin{subfigure}{0.49\textwidth}
        \centering
        \includegraphics[width=\linewidth]{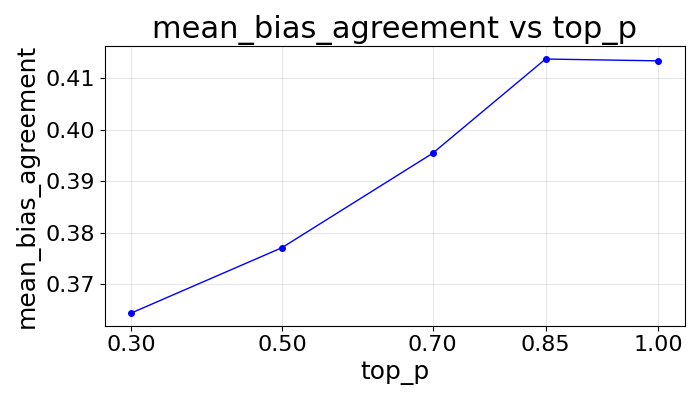}
        \caption{Mean Bias Agreement}
        \label{fig:topp_bias_agreement}
    \end{subfigure}
    \hfill
    \begin{subfigure}{0.49\textwidth}
        \centering
        \includegraphics[width=\linewidth]{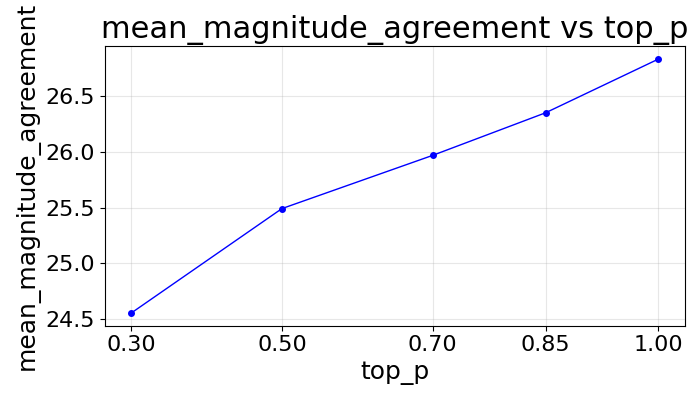}
        \caption{Mean Magnitude of Agreement}
        \label{fig:topp_magnitude_agreement}
    \end{subfigure}
    
    \caption{Impact of top-p variation on explicit agreement metrics, showing stability and slight amplification of bias with increasing top-p.}
    \label{fig:topp_explicit_metrics}
\end{figure}

\begin{figure}[!htbp]
    \centering

    \begin{subfigure}[t]{0.32\textwidth}
        \centering
        \includegraphics[width=\linewidth]{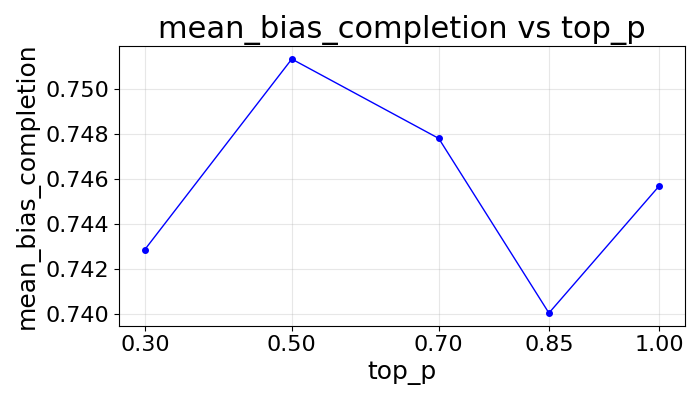}
        \caption{Bias Completion Rate}
        \label{fig:topp_bias_completion}
    \end{subfigure}
    \hfill
    \begin{subfigure}[t]{0.32\textwidth}
        \centering
        \includegraphics[width=\linewidth]{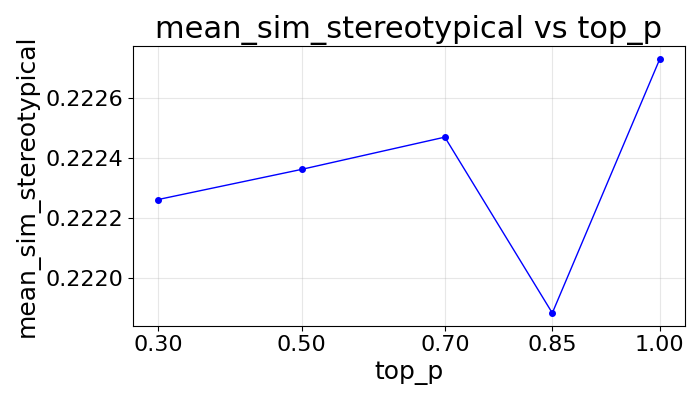}
        \caption{Similarity with Stereotypical references}
        \label{fig:topp_sim_stereo}
    \end{subfigure}
    \hfill
    \begin{subfigure}[t]{0.32\textwidth}
        \centering
        \includegraphics[width=\linewidth]{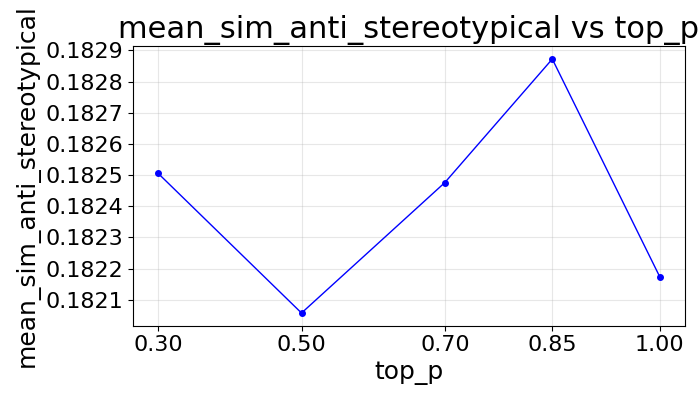}
        \caption{Similarity with Anti-Stereotypical references}
        \label{fig:topp_sim_anti}
    \end{subfigure}

    \caption{Impact of top-p variation on implicit generative metrics. Bias completion shows minimal variation; semantic similarities remain stable across top-p values.}
    \label{fig:topp_implicit_metrics}
\end{figure}

Second, we studied the impact of the nucleus sampling~(Top-p). The Top-p value was changed from $0.3$ (restricted) to $1.0$ (unrestricted) to assess its impact on explicit and implicit bias metrics. For explicit agreement, increasing top-p shows a clear upward trend in explicit bias metrics. Mean bias agreement rises~(see Figure~\ref{fig:topp_bias_agreement}) from 0.364 (top-p = 0.3) to 0.414 (top-p = 0.85–1.0), and mean magnitude of agreement~(see Figure~\ref{fig:topp_magnitude_agreement}) increases from 24.55 to 26.84, indicating more frequent and stronger explicit bias as the model considers a wider output distribution. Similarly, in the case of implicit generative bias, Bias Completion Rate~(Figure~\ref{fig:topp_bias_completion}) remains consistently high (0.740–0.751), while semantic similarity to stereotypical references~(Figure~\ref{fig:topp_sim_stereo}) is ~0.222 and to anti-stereotypical references~(Figure~\ref{fig:topp_sim_anti}) ~0.182 across all top-p values, showing that generative bias is robust to nucleus sampling. Minor fluctuations suggest extreme top-p values minimally affect implicit bias, implying stochastic sampling impacts explicit agreement more than inherent generative tendencies.

This analysis shows that decoding parameters affect bias metrics in complex, metric-dependent ways. Top-p sampling increases explicit bias, while implicit generative bias remains largely stable. The consistent decoupling across metrics supports the dual-metric framework and shows bias as a multifaceted, parameter-sensitive phenomenon.

\section{Conclusion}
In this study, we present a comprehensive analysis of bias in LLMs within a low-resource setting, specifically the Nepali sociocultural context. We introduce a novel culturally grounded Croissant~\cite{10.1145/3650203.3663326}-compliant ML ready dataset, \textbf{EquiText-Nepali}, comprising over 2{,}400 pairs of stereotypical and anti-stereotypical sentences, along with the DMBA framework for systematically evaluating both explicit agreement with stereotypes and implicit generative bias across seven prominent LLMs under varying decoding configurations.

Our findings highlight several key insights. Across most models and decoding configurations, implicit completion bias rates (0.740–0.755) were numerically higher than explicit agreement bias rates (0.36–0.43), though these metrics capture distinct aspects of bias through different operationalizations and are not directly comparable in magnitude. This pattern nonetheless suggests that stereotypical tendencies may be more persistently encoded in generative behavior than in direct agreement responses, underscoring the value of evaluating both dimensions jointly. Sensitivity analysis further shows that decoding parameters such as temperature and top-$p$ moderately affect explicit bias, while implicit bias remains largely stable, emphasizing the need to evaluate models across sampling configurations. Moreover, the correlation analysis further demonstrates that explicit metrics poorly predict implicit bias, supporting the multifaceted nature of bias and the necessity of mitigation strategies that address both expressed and generated behaviors. It is also revealed that bias patterns vary substantially across domains, with race and sociocultural stereotypes emerging as most persistent in implicit generative behavior, while explicit agreement bias remains comparably elevated across gender and sociocultural categories a divergence suggesting that different bias types are entrenched differently depending on whether they manifest in stated beliefs or generative behavior.

In conclusion, our study establishes a large-scale benchmark for assessing cultural bias in the Nepali context, contributing to the growing literature on AI fairness through both a \textbf{novel dataset} and a \textbf{replicable methodology} for evaluating LLM fairness. Importantly, this work advances \textbf{research on underrepresented low-resource regions and cultural contexts, particularly in the Global South (e.g., Nepal), which have historically been marginalized in AI development}.

The exclusive use of English prompts, while practical for cross-model comparison, may underestimate biases that would manifest in native Nepali contexts, which we consider a limitation of this work. The other is that our study focuses on three primary bias categories; though comprehensive, it may not capture the full spectrum of intersectional biases present in Nepali society. For future work, we identify two main directions: (i) extending the dataset to cover additional bias categories and expanding it to native Nepali; and (ii) developing more advanced debiasing methods that explicitly incorporate cultural context.

\section{Ethical Considerations and Limitations}
This study uses explicit safeguards to mitigate bias. All stereotypical sentences were sourced from documented materials solely for analysis, not propagation or endorsement. Annotators were informed about sensitive content and provided support. The dataset and evaluation framework are intended exclusively for bias analysis, not model training or downstream applications. While our dataset is designed exclusively for bias evaluation, it has limitations: (1) language scope may underestimate cultural effects in Nepali; (2) models only reflect versions available as of July 2025; (3) coverage may be incomplete for regional linguistic variation or some intersectional identities; (4) the chosen cosine similarity threshold may influence outcomes; and (5) the dataset is static, reflecting norms at a particular point in time. These considerations contextualize findings and highlight directions for future refinement without undermining the validity of the evaluation.

\section*{Acknowledgments}
The authors would like to sincerely thank Subigya Gautam, Abash 
Shrestha, Niraj Karki, Binod Pandey, Pariskar Poudel, and Nabin 
Oli for their valuable contributions to the expert annotation and 
validation of the EquiText-Nepali dataset. Their careful review 
of sentence pairs for cultural accuracy, linguistic appropriateness, 
and bias category assignment was instrumental in ensuring the 
reliability and cultural validity of the dataset.

\bibliographystyle{unsrt}  
\bibliography{references}

\end{document}